\def\year{2021}\relax
\documentclass[letterpaper]{article} 
\usepackage{aaai21}  
\usepackage{times}  
\usepackage{helvet} 
\usepackage{courier}  
\usepackage[hyphens]{url}  
\usepackage{graphicx} 
\usepackage{natbib}  
\usepackage{caption} 

\usepackage[export]{adjustbox}
\usepackage{amsmath,amssymb}
\usepackage{multirow}
\usepackage{algorithm}
\usepackage{algpseudocode}
\usepackage{blindtext}
\usepackage{bm}
\usepackage{subfig}
\usepackage{diagbox}

\usepackage{booktabs}

\usepackage[capitalise]{cleveref}
\usepackage[utf8]{inputenc}
\usepackage[english]{babel}
\usepackage[autostyle]{csquotes}
\usepackage[switch]{lineno}

\nocopyright

\usepackage{amsmath,amsfonts,bm}

\def\eqref#1{equation~\ref{#1}}

\def\1{\bm{1}}

\def\mT{{\bm{T}}}

\def\mW{{\bm{W}}}

\DeclareMathAlphabet{\mathsfit}{\encodingdefault}{\sfdefault}{m}{sl}
\SetMathAlphabet{\mathsfit}{bold}{\encodingdefault}{\sfdefault}{bx}{n}
\newcommand{\tens}[1]{\bm{\mathsfit{#1}}}

\def\tT{{\tens{T}}}

\def\tW{{\tens{W}}}
\def\tX{{\tens{X}}}

\newcommand{\E}{\mathbb{E}}

\newcommand{\R}{\mathbb{R}}

\newcommand{\normlone}{L^1}

\newcommand{\normlp}{L^p}

\DeclareMathOperator*{\argmax}{arg\,max}
\DeclareMathOperator*{\argmin}{arg\,min}

\title{A Progressive Sub-Network Searching Framework for Dynamic Inference}

\author{
    Li Yang,
    Zhezhi He,
    Yu Cao,
    Deliang Fan
    \\
}
\affiliations{School of Electrical, Computer and Energy Engineering, Arizona State University, Tempe, AZ\\ \{lyang166, zhezhihe, ycao, dfan\}@asu.edu}

\begin{document}
\maketitle

\begin{abstract}
Many techniques have been developed, such as model compression, to make Deep Neural Networks (DNNs) inference more efficiently. Nevertheless, DNNs still lack excellent run-time dynamic inference capability to enable users trade-off accuracy and computation complexity (i.e., latency on target hardware) after model deployment, based on dynamic requirements and environments. 
Such research direction recently draws great attention, where one realization is to train the target DNN through a multiple-term objective function, which consists of cross-entropy terms from multiple sub-nets. Our investigation in this work show that the performance of dynamic inference highly relies on the quality of sub-net sampling. 
With objective to construct a dynamic DNN and search multiple high quality sub-nets with minimal searching cost, we propose a progressive sub-net searching framework, which is embedded with several effective techniques, including trainable noise ranking, channel group and fine-tuning threshold setting, sub-nets re-selection. 
The proposed framework empowers the target DNN with better dynamic inference capability, which outperforms prior works on both CIFAR-10 and ImageNet dataset via comprehensive experiments on different network structures. Taken ResNet18 as an example, our proposed method achieves much better dynamic inference accuracy compared with prior popular Universally-Slimmable-Network by 4.4\%-maximally and 2.3\%-averagely in ImageNet dataset with the same model size.
\end{abstract}

\section{Introduction}

Recently, Deep Neural Networks (DNNs) grow into more complex structures consisting of deeper layers, larger model size, and denser connections. 
Such \textquote{bulky} models rise challenges to their hardware deployment, for both edge- and cloud-computing systems. 
The most common solution is to compress the target DNN for resource-efficient deployment, in terms of latency, throughput and etc. ~\cite{alvarez2016learning,wen2016learning,li2016pruning,liu2017learning,he2019filter}. 
Consequently, it normally leads to a static/fixed model that is not capable of adjusting or re-configuring its computation complexity (i.e., inference structure, latency) to the dynamic available resource or environment constraint, in run-time after model deployment.

To address the static inference issue, dynamic DNN is proposed with empowered dynamic inference capability. One approach is the input-dependent dynamic DNN~\cite{liu2018dynamic}, where a sub-net is selected as the inference path on-the-fly w.r.t input.
Such input-dependent sub-net selection can be achieved via the controller module~\cite{liu2018dynamic, wu2018blockdrop, wang2018skipnet}, or inserting a cascade of classifiers operating on the features of internal layers~\cite{huang2017multi}.
However, the input-dependent dynamic inference owns an static hardware utilization on average, which still not meets the expectation of dynamic hardware utilization. 
As the countermeasure, Yu et. al. propose the Slimmable neural Network (S-Net)~\cite{yu2018slimmable} and its optimized counterpart (US-Net)~\cite{yu2019universally}, which can switch the inference structure among the predefined sub-net candidates in an input-independent and run-time fashion, to dynamically trade-off inference accuracy and computation complexity.
Note that, the predefined sub-nets in S-Net/US-Net~\cite{yu2018slimmable,yu2019universally} are naively sampled by multiplying the original channel-width w.r.t a multiplier. Such channel-width multiplier is \textit{uniformly} applied on all layers (e.g., convolution/fully-connected layers) throughout the entire target DNN. 
If viewing the sub-net searching as a DNN pruning processing, prior works~\cite{li2016pruning,molchanov2017variational,wen2016learning,zoph2016neural,liu2018progressive,liu2018darts} have discussed that, the parametric layers own non-identical sensitivity (i.e., non-uniform) to downscale of their weight tensor size. 
It reveals the potential to improve the accuracy of S-Net/US-Net~\cite{yu2018slimmable,yu2019universally} based dynamic DNN via better sub-net searching strategy, instead of naive uniform counterpart.
Thus, in this work, we focus on investigating: \textit{How to perform the non-uniform sub-net searching for optimal dynamic inference?}

To sample multiple non-uniform sub-nets efficiently from a given DNN as super-network, we propose to leverage the progressive neural network pruning as the backbone technique of sub-net searching. 
The progressive pruning~\cite{yang2018netadapt} is one well-known model compression technique, which iteratively drops unimportant weights (e.g., one channel/iteration~\cite{yang2018netadapt}) to progressively shrink model size, without hampering the accuracy of initial super-net model.
The progressive pruning fits well for sub-net searching to construct dynamic DNN, due to the following properties: 
1) Massive sub-net candidates could be identified through progressive pruning; 
2) The weights of a sampled smaller sub-net $\mathcal{A}$ is always the subset of its larger counterpart $\mathcal{B}$ (i.e., $\tW_{\mathcal{A}} \in \tW_{\mathcal{B}}$). Such property is vital to maximize the accuracy of target dynamic neural network, which will be discussed later.

Nevertheless, there still remains several challenges to overcome while incorporating the progressive pruning as the progressive non-uniform sub-net searching for dynamic neural network. It could be summarized as:
\textbf{1) Searching Quality:} the vanilla progressive pruning normally adopts the $\normlp$-norm of weights ($||\tW||_1$ or $||\tW||_2$) as the importance criterion to drop weights, which can be further optimized for identifying sub-nets with higher quality;
\textbf{2) Searching Speed:} the searching cost of progressive sub-net searching is very high, thus countering the \textit{scalability} issue (e.g., 35.8 GPU hours for ResNet20 on small CIFAR-10 dataset).
First, to enhance searching quality, we propose a new method through injecting channel-wise zero-mean Gaussian noise with trainable variance upon weight during training, then leveraging such trained variance as weight importance criterion (named as \textit{trainable noise ranking}) for progressive sub-net searching. 
Second, to further boost searching speed, we also enlarge the granularity per searching iteration from single channel to multiple channels as a group, combined with optimized fine-tuning configurations. 
In summary, our technical contributions include:
\begin{itemize}

    \item Given a target DNN, we empower it the state-of-the-art dynamic inference capability which can dynamically trade-off accuracy and computation complexity (i.e., latency on target hardware platform equivalently) on-the-fly via switching among multiple sampled sub-nets.
    
    \item A progressive sub-net searching framework is proposed to quickly identify high-quality sub-nets. A series of novel techniques are developed and utilized as well under this framework, e.g., trainable noise ranking, channel group and fine-tuning threshold setting, and etc.
    
    \item Taken the classic object classification as a study case, the framework outperforms prior works via comprehensive experiments on both CIFAR-10 and ImageNet dataset on general-purpose computing devices (i.e., CPU and GPU). 
    
    \item Heuristic discussion with experiments are provided to explore optimal sub-net structures for different networks, e.g., MobileNet, ResNet, AlexNet and VGG.
\end{itemize}


\begin{figure*}[ht]
  \centering
  \includegraphics[width=1.0\linewidth]{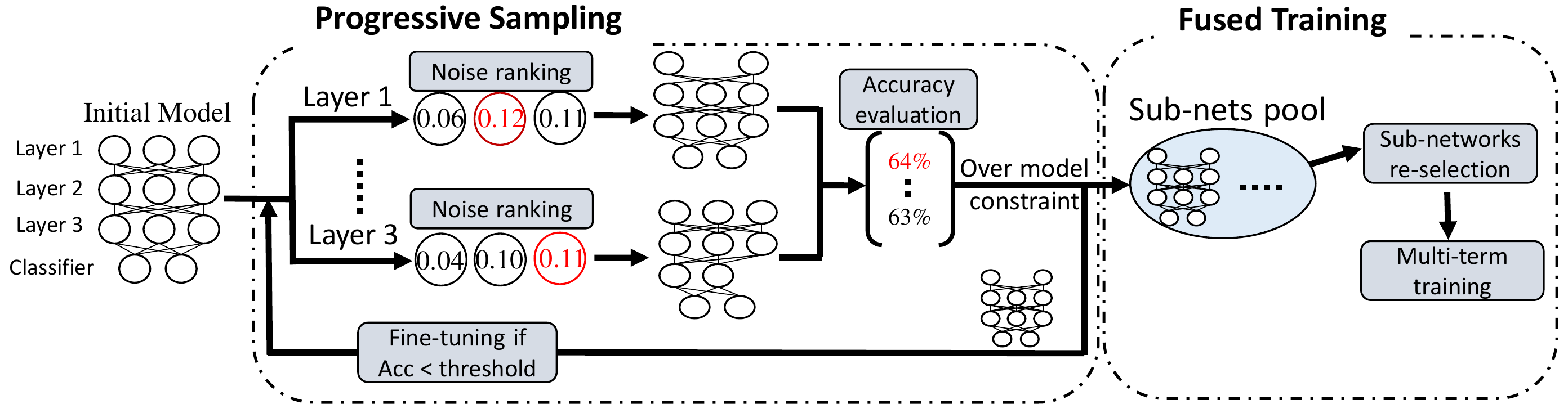}
\caption{
Flowchart of the progressive sub-nets searching for dynamic inference. 
First, we progressively search various sub-nets from the initial model with different model capacities. Note that, sub-nets are partially sharing the weights as indicated by the overlapped channel index. Then, through the followed fused training step, the initial model can act as a \textit{dynamic model}, while each sub-nets can perform inference independently at different power, speed, accuracy. 
}
\label{fig:method_overflow}
\end{figure*}

\section{Related Works and Background}
\label{sec:related_work}

\subsection{Dynamic Neural Network}
\label{sec:dy_infer}

As discussed above, dynamic DNN with the capability of switching inference structures has been studied in prior works~\cite{huang2017multi,liu2018dynamic,wang2018skipnet}. They have studied the input-dependent dynamic DNNs which change the inference structure per input sample. As such input-dependent DNN cannot benefit inference in real hardware implementation, it is out of our scope. 

Here we focus on explaining the Slimmable neural Network (S-Net)~\cite{yu2018slimmable}, which is a dynamic DNN that user can switch the inference structure among $N$ pre-defined sub-nets ($N=4$ in~\cite{yu2018slimmable}). In S-Net, each sub-net owns different computation complexity and accuracy. Generally, the accuracy of sub-net is proportional to its computation complexity. The method adopted by S-Net is quite straight-forward and described as follow:

\textbf{1) Uniform sub-nets generation.} S-Net first generates $N=4$ sub-nets from the initial full-size model (aka. super-network). Each $i$-indexed ($i\in\{1,2,3,4\}$) sub-net is acquired by uniformly applying a multiplier $m_i\in\{0.25, 0.5,0.75,1.0\}$ on the input/output channels of entire super-network. 
Taken one fully-connected layer in the super-network with weight matrix $\mW \in \R^{p^2}$ as an example, the weight matrices in this layer of four sub-nets $\{\mW_1, \mW_2, \mW_3, \mW_4\}$ are in shape of:
\begin{equation}
\label{eq:uniform_sub-nets}
\begin{gathered}
    \{ \R^{(0.25\times p)^2}, \R^{(0.5\times p)^2}, \R^{(0.75\times p)^2}, \R^{(1.0\times p)^2} \} \\
    s.t.~~\mW_1 \subseteq \mW_2 \subseteq \mW_3 \subseteq \mW_4
\end{gathered}    
\end{equation}
where $p$ is the input and output size. For simplicity, the weight tensor sets of $i$-th sub-nets is denoted by $\{\tW_i\}$.

\textbf{2) Multi-objective training.} To enable S-Net switch among sub-nets with un-compromised individually inference accuracy, S-Net trains the target DNN by conventional back-propagation with a multi-term objective function, which can be expressed as:
\begin{equation}
\label{eqt:mul_term}
    \min~\E_{\tX}\Big{(}\textstyle\sum_{i=1}^{N}\mathcal{L}(f(\tX, \{\tW_i\} );\mT) \Big{)}
\end{equation}
where $\tX$ is the mini-batch of inputs with corresponding targets$\mT$. $\mathcal{L}(\cdot;\cdot)$ calculates the cross-entropy loss of DNN output and target. $f(\tX,\{\tW_i\})$ computes the output of sub-net parameterized by $\{\tW_i\}$.

Through the above two sequential steps, with single DNN as super-network, S-Net can switch among its sub-nets on-the-fly. To further improve S-Net performance, Yu et al. also propose several techniques in their extended work called Universally Slimmable Networks (US-Net)~\cite{yu2019universally}, including in-place distillation, post-statistics of batch normalization and etc.

\subsection{Progressive Model Pruning}
\label{sec:pro_pruning}
Model pruning~\cite{han2015learning} is an important technique in DNN model compression, where the pruned model with reduced model size can run in the hardware with less computation workload. 
Note that, all the pruning method discussed hereafter are structured pruning, where the basic weight group to be dropped is the entire output channel for convolution and fully-connected layer.
Prior pruning works can be divided into two categories: Rule- and Progressive-based. 

The rule-based pruning methods~\cite{alvarez2016learning,wen2016learning,li2016pruning,liu2017learning,he2019filter} normally apply a weight penalty term (e.g., group Lasso~\cite{wen2016learning,yang2019harmonious}) in objective function or differentiable masking function on weights, during training process.
Progressive pruning~\cite{yang2018netadapt} usually takes a pretrained model as initialization, then gradually shrinking the full size model, which can generate a family of simplified sub-nets with different model sizes.
Given a target DNN with $L$ layers and its $l$-th layer ($l\in{1,2,..,L}$) has $p_l$ output channels, the weight tensors of entire DNN $\{\tW_l\}_{l=1}^L$ can be re-factorized into $\{\tW_j\}_{j=1}^{\sum p_l}$. For each iteration of progressive pruning, it attempts to zero-out one channel of weights $\tW_k, k\in\{1,2,..., \sum p_l\}$, while minimizing the potential accuracy degradation. Such process can be described as:
\begin{equation}
\label{eqt:pro_def}
    \argmin_k ACC_{\textrm{val}}(f(\tX, \{\tW_j\}_{j=1}^{\sum p_l} \setminus \tW_k );\tT)
\end{equation}
where $\setminus$ denotes the element exclusion from set. As the method adopts brute-force to prune the target DNN channel-by-channel, the computation cost is $\sum p_l(\sum p_{l}+1)/2$.

Both pruning methods will lead to a non-uniformly pruned DNN.
However, in comparison to the rule-based pruning methods, progressive pruning owns the following merits: 1) It can produce multiple pruned sub-nets since the full size super-net is progressively shrunk; 
2) the smaller pruned network $\mathcal{A}$ is certainly the subset of its larger counterpart $\mathcal{B}$ (i.e., $\tW_{\mathcal{A}} \in \tW_{\mathcal{B}}$).


\section{Progressive Sub-Network Searching}


In this work, we aim to search multiple high quality sub-nets with minimized searching cost for dynamic inference. 
We propose different novel techniques to overcome below three challenges in this process:
\begin{enumerate}
    \item How to search \textbf{multiple} sub-nets?
    \item How to improve the \textbf{quality} of identified sub-nets?
    \item How to improve the searching \textbf{speed}, while maintaining the sub-net performance?
\end{enumerate}

First, we propose to leverage the progressive pruning as the backbone technique of sub-net searching, as it can produce multiple sub-nets while the smaller pruned network $\mathcal{A}$ is the subset of its larger counterpart $\mathcal{B}$ (i.e., $\tW_{\mathcal{A}} \in \tW_{\mathcal{B}}$). This property is vital to maximize the accuracy of target dynamic neural network, since the weights of all sub-nets are partially shared during the training for dynamic inference. However, adopting brute-force method to prune the target DNN channel-by-channel is computationally prohibitive (e.g., 35.8 GPU hours for ResNet20 on small CIFAR-10 dataset).   
Thus, to search sub-nets more efficiently, instead of brute-force pruning, the weight-norm ($||\tW||_1$ or $||\tW||_2$) is normally adopted as channel importance ranking method to determine the order of channel pruning in a layer. That it, only the channel with small weight-norm value in a layer will be pruned. By doing this, $L$ times pruning is needed in each iteration given a target DNN with $L$ layers, and the computation cost can be reduced to $L^2$. However, weight-norm ranking determines the quality of sampled sub-nets.

Second, although sub-net searching by channel importance ranking can reduce computation cost, it also influences the quality of sub-nets. To improve the quality of sub-nets, we propose a novel training method by injecting channel-wise zero-mean Gaussian noise with trainable variance upon weights, while leveraging such trained variance as weight importance criterion (named as \textit{trainable noise ranking}) for progressive sub-net searching.
Furthermore, in terms of dynamic inference, we further propose \textit{sub-net re-selection} after progressive searching to exclude low-quality (or find optimal) sub-nets by inter-iteration comparison. 

Third, to further boost the searching speed while maintaining sub-net accuracy, we propose to adopt two optimization techniques: 1) enlarging the granularity per searching iteration from single channel to multiple channels as a group; 2) optimizing fine-tuning configurations. 



\cref{fig:method_overflow} illustrates the overflow of the proposed method. It can be divided into two successive steps:
\begin{enumerate}
    \item In the first step, we progressively sample non-uniform sub-nets with different model sizes. In each iteration, we evaluate the accuracy of each candidate sub-net, which is pruned layer-by-layer. In addition, the proposed \textit{trainable noise ranking} is utilized to determine the priority of channels to be pruned in each layer. Then the sub-nets with maximum accuracy as mentioned in~\cref{eqt:pro_def} is selected and feed into sub-net pool.

    \item In the second step, we further process sub-net pool to guarantee that the evaluated accuracy of larger sub-net is always higher than smaller sub-nets. Then, the initial model which includes these re-selected sub-nets is trained via an ensemble loss for multiple objective optimization as expressed in~\cref{eqt:mul_term}.  
    Note that, all sub-nets partially share the weights of initial model as indicated by the overlapped channel index.
    Finally, the trained model can act as a dynamic model whose sub-nets can perform inference independently at different power, speed, accuracy.
\end{enumerate}

\subsection{Trainable Noise Ranking}
\label{sec:trainable_noise_ranking}


Training network with weight noise injection is an effective technique to perform model regularization, thus improving model robustness against input variation~\cite{liu2018towards,he2019parametric}. But different from aiming to improve model robustness, we are \textit{the first to propose that such trainable weight noise can also be used to sub-net sampling or pruning from a given DNN}. In practice, we introduce the channel-wise Gaussian noise to the pre-trained model for both convolutional and fully connected layers, and then the trainable noise variance (i.e. magnitude) is used for channel importance ranking in sub-net sampling. The weight noise injection can be mathematically described as:
\begin{equation}
\label{eqt:PNI_definition}
    \Tilde{w}_{l,i} = w_{l,i} + \beta_{l,i} \cdot \eta_{l,i}; \quad \eta_{l,i} \sim \mathcal{N}(0, \sigma_{w_{l}}^2)
\end{equation}
where $w_{l,i}$ is the $i$-th channel of noise-free weight $\bm{w}_l$ in $l$-th layer. $\eta_{l,i}$ is the noise term samples from Gaussian distribution with zero mean. It share the same variance $\sigma_{w_l}^2$ of the weight $\bm{w}_l$ in training. $\beta_{l,i}$ is the channel-wise noise variance that scales the magnitude of injected noise $\eta_l$. 
Note that, we adopt the scheme that $\eta_l$ shares the identical weight variance with $\bm{w}_l$ as in~\cref{eqt:PNI_definition}, thus the injected additive noise relies on $\beta_l$ and the distribution of $\bm{w}_l$ simultaneously. 

Utilizing noise variance $\bm{\beta}$ to prune weight channels is guided by the following two properties: 1) it is a channel-wise parameter that can be automatically updated according to current weight distribution during training. Thus, the noise variance $\bm{\beta}$ is different among channels. 2) The impact of $\bm{\beta}$ is to scale the magnitude of corresponding noise, which represents the strength of noise. For example, the weight channel with larger value of $\bm{\beta}$ means that stronger regularization is needed to keep robustness and accuracy. It implies that this weight channel is not as important as the one with smaller trained noise magnitude. 
So we conjecture that 
\textit{"the larger value of coefficient $\beta_{l,i}$ represents the corresponding weight channel is less important."}
Based on this hypothesis, we apply the channel-wise $\bm{\beta}$ to guide sub-net searching. As shown in~\cref{fig:method_overflow}, the weight channel with larger $\beta_{l,i}$ is pruned first in each layer. The progressive searching via trainable noise ranking can be re-formalized as:
\begin{equation}
\begin{gathered}
\label{eqt:pro_noise}
    \argmin_{k^\star} ACC_{\textrm{val}}(f(\tX, \{\tW_j\}_{j=1}^{\sum p_l} \setminus \tW_{k^\star} );\tT) \\
    \textrm{s.t.}~ k^\star = \argmax_{k} \beta_{j,k}
\end{gathered}
\end{equation}
Note that, we introduce~\cref{eqt:PNI_definition} to the pre-trained model and then only the parameter $\bm{\beta}$ is used to select the channel to be pruned. Our experiments show that the noise ranking can achieve the same or even better results than typical norm-based ranking, especially on larger network. 

\subsection{Channel Group and Fine-tuning Threshold}
\label{sec:channel_group_and_fine_tuning}
\begin{figure}[t]
  \centering
  \includegraphics[width=0.9
  \linewidth]{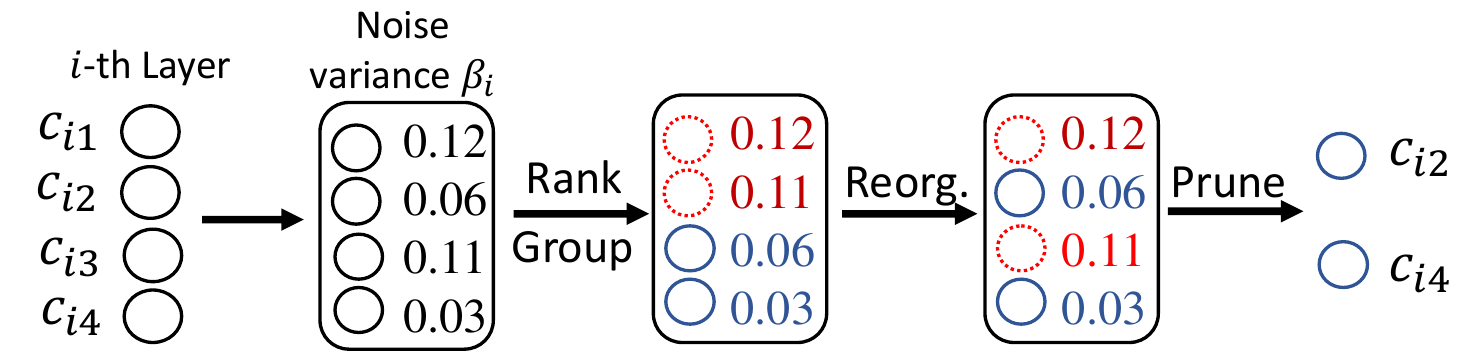}
  \caption{Example of channel group setting. 4 weight channels in a layer are assigned into 2 channel groups. Thus, 2 channels are pruned in each searching iteration}
\label{fig:ch_group}
\end{figure}

To further reduce the searching space, we enlarge the searching granularity per iteration from single channel to multiple channels as \textit{channel group}. 
\cref{fig:ch_group} gives an example to explain the flow of channel group setting.
First, we rank the 4 weight channels via trainable noise variance $\beta_l$, and combine the adjacent 2 channels to be one group. Then one group will be pruned in each searching iteration, instead of one channel. For generalization, we name the group $G$ as the total number of groups per layer, which could be the power of 2 (e.g., 4, 8, 16) for efficient computing in hardware, and is identical for all layers. Then, given a target DNN with $L$ layers and $G$ groups per layer, the \textit{searching cost is reduced to $L\times G$}. 

Moreover, further fine-tuning is needed for sub-net searching to optimize weights and maximize the evaluated accuracy of current sampled sub-net. Otherwise, the trainable noise ranking may give a non-optimal decision according to accuracy metric, as the ranking factor $\bm{\beta}$ w.r.t the \textquote{old} weights of full size model.
However, fine-tuning is time-consuming especially for a large dataset (e.g., ImageNet). Inspired by the model pruning~\cite{han2015learning} that reveals DNN model redundancy, we find that fine-tuning is not necessary for larger sampled sub-nets. Thus, we try to speedup the searching by minimizing the fine-tuning time cost. To do so, we set an accuracy target as a threshold, where the sampled sub-nets will only be fine-tuned when its accuracy is lower than that. Our experiments indicate that such threshold setting can further reduce the sampling cost without influencing the performance of dynamic inference, which are elaborated in the experiment section.  

\subsection{Sub-Nets Re-Selection and Fused Training}
\label{sec:select}

After progressive searching, all pruned sub-nets are fed into a sub-net pool. These sub-nets are sampled via~\cref{eqt:pro_noise} gradually, which can be considered as \textquote{optimal} structures intra-iterations, but lacking of comparing inter-iterations. Due to model redundancy, the accuracy of  smaller sub-nets $\mathcal{A}$ may be higher than a larger sub-net $\mathcal{B}$ (i.e., $\tW_{\mathcal{A}} \in \tW_{\mathcal{B}}$, but $ACC_{\textrm{val}}(\tW_{\mathcal{A}}) > ACC_{\textrm{val}}(\tW_{\mathcal{B}})$). As the countermeasure, we add a constraint to remove the larger sub-nets with lower accuracy w.r.t smaller counterpart (e.g. $\tW_{\mathcal{B}}$). The constraint can be formulated as:
\begin{equation}
    \small{ACC_{\textrm{val}}(f(\tX, \tW_{\{1,..,i-1\}}));\tT) >    ACC_{\textrm{val}}(f(\tX, \tW_{i}));\tT)}
\end{equation}
where $i$ is the sub-nets index.
Then after sub-net re-selection, the rest sub-nets will be applied to fused training by using multiple-term objective optimization for dynamic inference as mentioned in \cref{eqt:mul_term}. 

\begin{figure*}[t]
\centering
  \includegraphics[width=0.33 \linewidth]{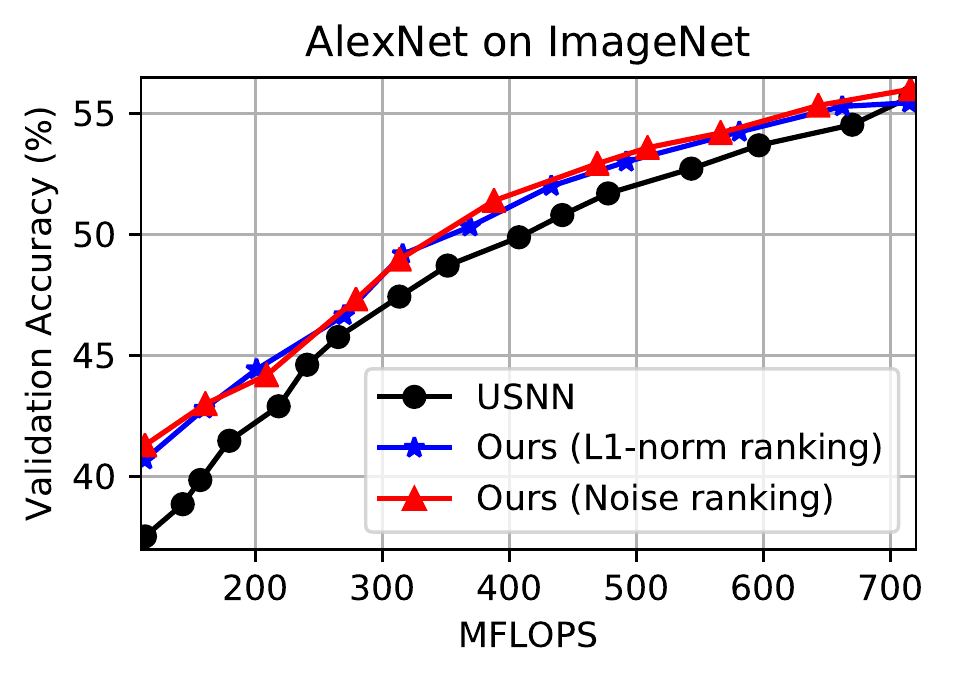}
  \includegraphics[width=0.33 \linewidth]{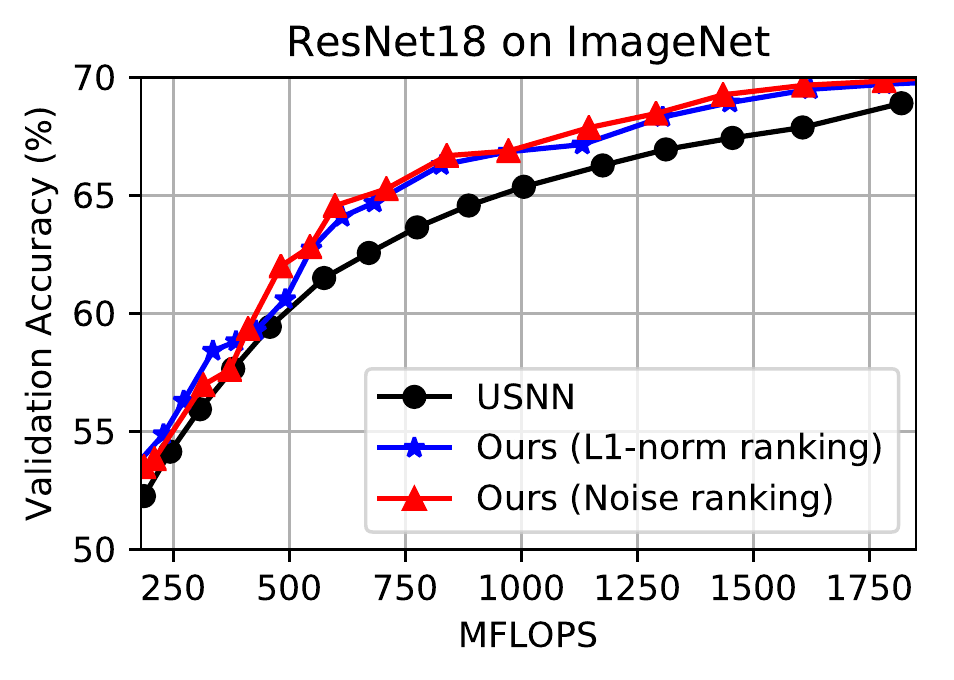}
  \includegraphics[width=0.33 \linewidth]{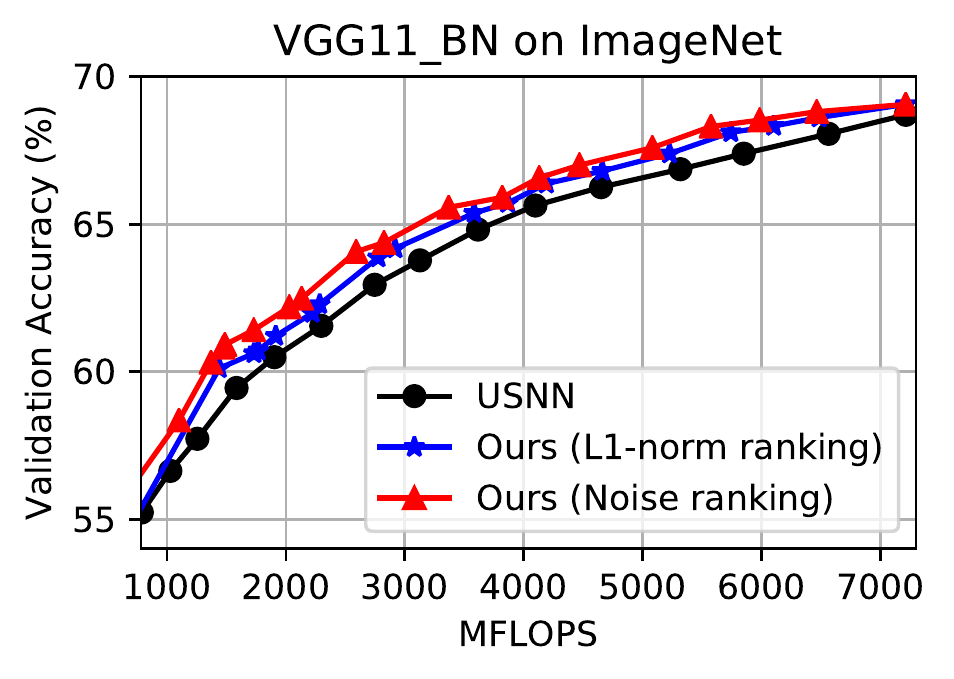}

\vspace{-1em}
\caption{Trade-off between accuracy and FLOPs for various networks on ImageNet. detailed numbers in Appendix-\cref{tab:main_result} }
\label{fig:main_exp}
\end{figure*}

\section{Experiments}
\label{sec:exper}

\subsection{Experiment Setup}
In this work, we use the classic image classification task to examine the performance of our proposed framework. Two datasets are used, which are CIFAR-10 \cite{krizhevsky2009learning} and ImageNet \cite{deng2009imagenet}. 
For CIFAR-10 dataset, we test our framework on three different networks: ResNet20~\cite{he2016deep}, MobileNetV1~\cite{howard2017mobilenets} and VGG11~\cite{simonyan2014very}. For ImageNet dataset, we test on ResNet18, AlexNet~\cite{krizhevsky2012imagenet} and VGG11-BN.

\subsubsection{Sub-Nets Searching}
For CIFAR-10, we randomly choose 5000 images from validation dataset to do validate. The accuracy threshold is set to be $50\%$ and the channel group is 4 for ResNet20 and MobiletNetv1, 8 for VGG11. For ImageNet, we validate the sub-net accuracy on 10000 random picked images from validation dataset. The accuracy threshold is $40\%$ and the channel group is 4 for ResNet18, and 8 for MobileNetV1 and VGG11-BN.


\subsubsection{Fused Training}
The minimum sub-net model size is constraint to $0.25\times$ of complete model. For ResNet18 on ImageNet dataset, we train the network using momentum SGD optimizer, where the initial learning rate is 0.1, then scaled by 0.1 at epoch 30, 60, 80 respectively. For AlexNet and VGG11-BN, we use the same configuration as~\cite{simon2016imagenet}, which choose momentum SGD optimizer, and the initial learning rate is 0.05 and 0.01 respectively, both scaled by linear decaying. 

\subsection{Main Results}

\subsubsection{Uniform vs Non-uniform}
\label{subsec:non_uni}
In this work, we use Floating-point Operations Per Second (FLOPS) to indicate the computing complexity of sub-nets. 
\cref{fig:main_exp} depicts the sub-net accuracy and FLOPS trade-off for different networks with dynamic inference on ImageNet. Our proposed dynamic network consists of multiple non-uniform sub-nets sampled through trained noise ranking based progressive search. We mainly compare it with the state-of-the-art US-Net~\cite{yu2019universally}, which contains uniform sub-nets. It is clear that the non-uniform sub-nets sampled from our method provide much better accuracy than the uniform sub-nets with various model sizes for all different network structures. 

\subsubsection{Trainable Noise Ranking vs $\normlone$-norm
Ranking} 
As discussed in the related work section, $\normlp$-norm ranking is also a potential metric used in searching non-uniform sub-nets for dynamic inference. To demonstrate the efficacy of our proposed trainable noise ranking, we also compare it with $\normlone$-norm ranking, as depicted in \cref{fig:main_exp} (quantitative results are tabulated in \textit{appendix-\cref{tab:main_result}}). From the detailed quantitative results, we observe noise ranking outperforms $\normlone$-norm ranking and US-Net in most cases. Note that, the model size $M$ for different networks follows: $M(\textrm{AlexNet})<M(\textrm{ResNet18})<M(\textrm{VGG11\_BN})$ on ImageNet. It is also intriguing to find that our trainable noise ranking method works much better in larger networks, like VGG, compared with the smaller counterparts. 

\subsubsection{Searching Cost Comparison}
One of the main objective of our proposed trainable noise ranking based sampling method is to reduce searching complexity, thus to speedup searching process, especially compared with traditional progressive searching based on brute-force pruning. 
\cref{tab:sample_cifar} lists the time required for traditional progressive searching, trainable noise ranking and $\normlone$-norm ranking on CIFAR-10 and ImageNet, respectively, on 4-way NVIDIA Titan-Xp GPUs.
It is noteworthy that other proposed techniques, such as channel group and accuracy threshold setting, are both applied to both ranking methods. So they have almost the same searching cost. It can be easily seen that traditional progressive searching methods takes significantly more time compared with our trainable noise ranking method. 

\begin{table}[h]
\caption{Searching cost for six networks under different configurations. Note that, we approximate the searching cost of traditional progressive searching method on ImageNet dataset by running a few iterations.}

\label{tab:sample_cifar}
\begin{adjustbox}{width=\columnwidth,center}
\begin{tabular}{cccccc}
\toprule
\multicolumn{1}{c}{\multirow{2}{*}{Network}} & \multicolumn{3}{c}{GPU-hours} & \multicolumn{2}{c}{\multirow{2}{*}{\begin{tabular}[c]{@{}c@{}}Group setting/\\ No. sub-nets\end{tabular}}} \\ \cline{2-4}
\multicolumn{1}{c}{} & \multicolumn{1}{c}{\begin{tabular}[c]{@{}c@{}}Traditional\\ progressive\end{tabular}} & \multicolumn{1}{c}{\begin{tabular}[c]{@{}c@{}}$\normlone$-norm\\ ranking\end{tabular}} & \multicolumn{1}{c}{\begin{tabular}[c]{@{}c@{}}Noise\\ ranking\end{tabular}} & \multicolumn{2}{c}{} \\ \midrule
ResNet20 & 35.8 & 0.23 & 0.26 & Group 4 / 57 &  \\ \hline
MobileNetv1 & 462.0 & 0.30 & 0.31 & Group 4 / 41 &  \\ \hline
VGG11 & 41.9 & 0.09 & 0.09 & Group 8 / 59 &  \\ \hline \hline
ResNet18 & $2.6\times10^4$ & 7.1 & 7.0 & \multicolumn{2}{l}{Group 4 / 50} \\ \hline
AlexNet & $1.2\times10^5$ & 7.2 & 7.2 & \multicolumn{2}{l}{Group 8 / 41} \\\hline
VGG11-BN & $9\times10^5$ & 10.9 & 10.7 & \multicolumn{2}{l}{Group 8 / 60} \\
\bottomrule
\end{tabular}
\end{adjustbox}
\end{table}

\subsection{Analysis and Ablation study}
\subsubsection{Impact of channel group setting}
\begin{figure}[ht]
  \centering
  \includegraphics[width=\linewidth]{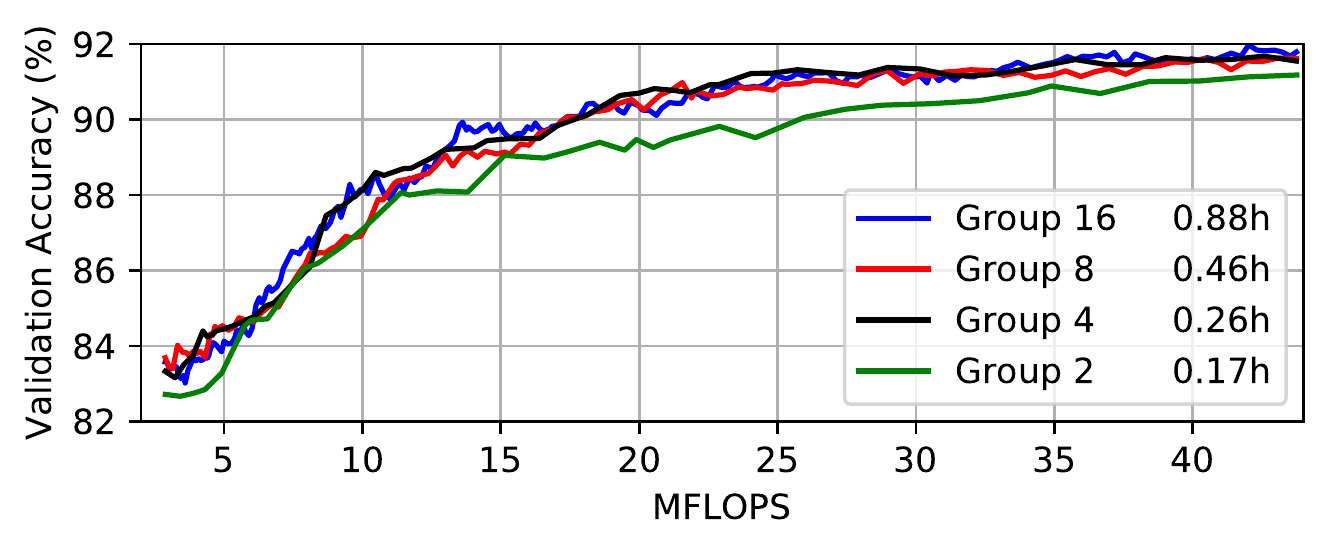}\\
\vspace{-1em}
\caption{Noise-ranking search with varying groups.}    
\vspace{-3em}
\label{fig:abs_group}
\end{figure}

Here we explore how the channel group setting will influence performance. Four different group settings with the corresponding searching costs are shown in~\cref{fig:abs_group}.
We define \textit{group $G$} as the number of groups in each layer. For example, group 4 represents the output channel of layer is divided into 4 groups. We observe that configuring the group size as 4, 8 or 16 achieves similar accuracy, in contrast to group size 2 in bad performance. Thus, we set group size as 4 or 8 for different networks as shown earlier.

\subsubsection{Accuracy threshold for fine-tuning}

\begin{figure}[ht]
\centering
\includegraphics[width=\linewidth]{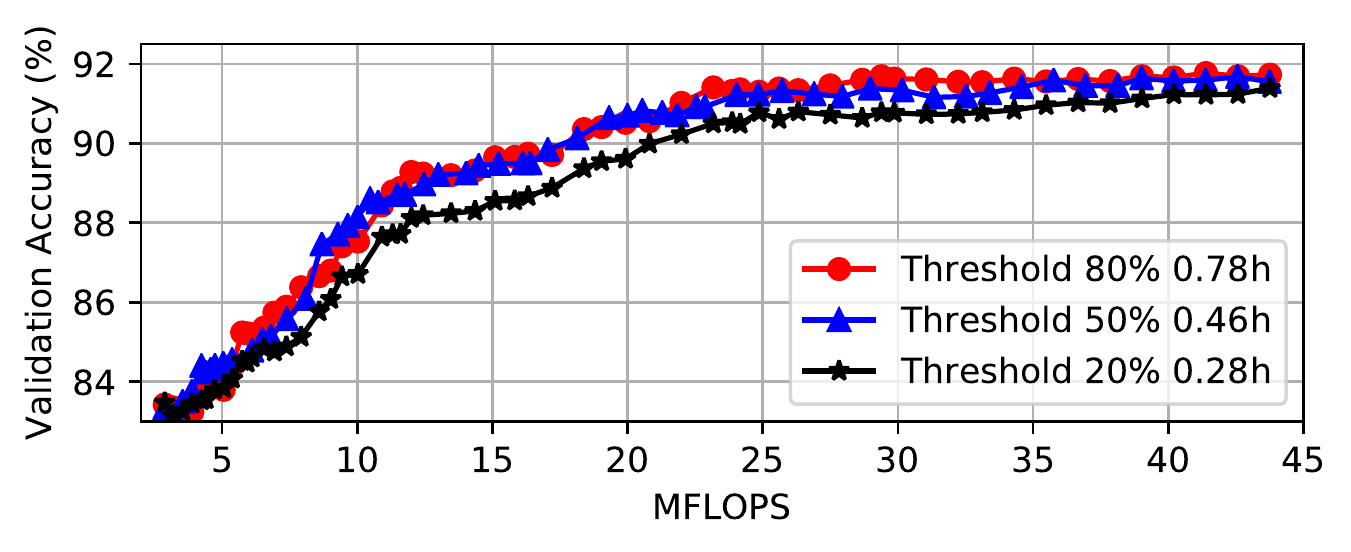} 
\vspace{-2em}
\caption{Noise-ranking search with varying accuracy threshold for fine-tuning, using ResNet20 on CIFAR10.}
\label{fig:thres}
\end{figure}

During the progressive sub-network search, one technique discussed earlier is that we only fine-tune the network if its accuracy is lower than a preset threshold. 
To demonstrate its effectiveness, we vary the fine-tuning accuracy threshold to be 80\%, 50\% and 20\% on CIFAR-10, as depicted in~\cref{fig:thres}. Considering the searching time cost, it is obvious that a higher threshold will require more fine-tuning and thus higher time cost during searching.
Moreover, the accuracy thresholds of 80\% and 50\% lead to similar accuracy versus FLOPS trade-off, compared to the threshold of 20\%. 
It indicates that searching with intermittent sub-nets fine-tuning is beneficial to identify better sub-nets, at the cost of extra computations.


\subsubsection{Sub-net re-selection}

\begin{figure}[ht]
\centering
\includegraphics[width=\linewidth]{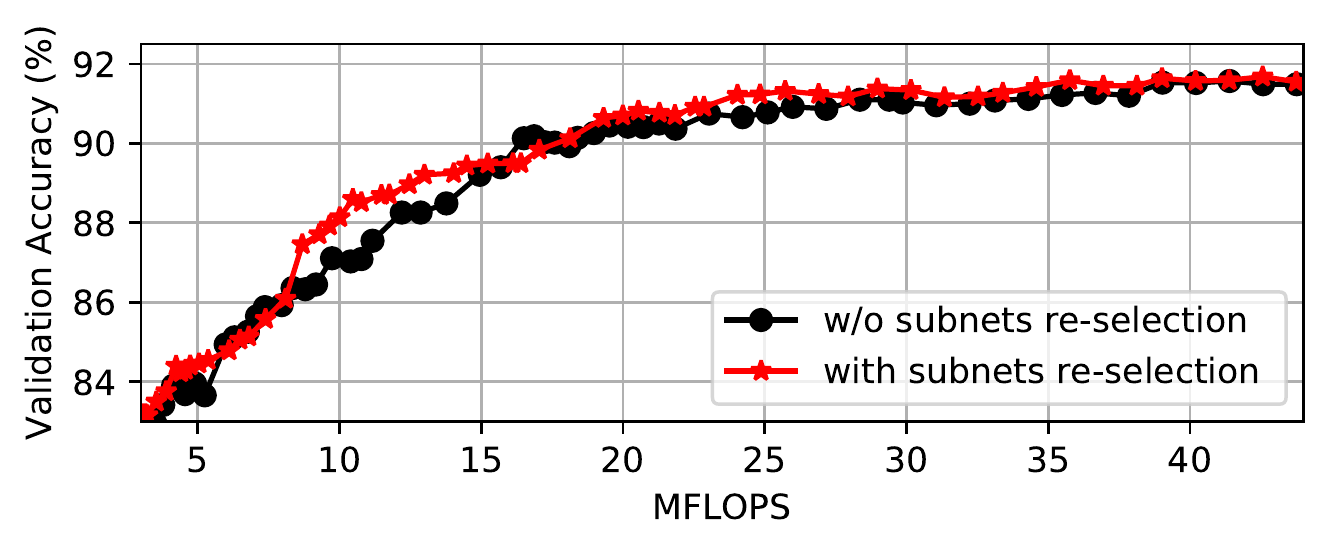}
\vspace{-2em}
\caption{Trade-off between accuracy and FLOPs, with and without sub-net re-selection on ResNet20.}
\vspace{-2em}
\label{fig:select}
\end{figure}

We adopt a technique called \textit{sub-net re-selection} to guarantee that a larger sub-net always has higher accuracy than a smaller sub-net in the pool. To demonstrate its effect on the overall performance, we conduct experiments to construct dynamic inference with and without sub-net re-selection, as shown in~\cref{fig:select}. It clearly shows that, with sub-net re-selection, we could get better accuracy for all sub-nets with different FLOPS, as well as eliminating the cases where larger network has smaller accuracy. We believe this technique is critical since weights are partially shared between sub-nets during the fused training for dynamic inference. If a non-optimal sub-net exists, it will influence the overall performance.

\subsubsection{Non-uniform structure of sampled sub-nets }
Our experiments have shown that our proposed non-uniform sub-net sampling provides better accuracy than the uniform counterpart with identical model size. It reveals that the layer-wise sensitivities over accuracy are different, which aligns with many prior network pruning works using different pruning methods. 
However, there is no standard golden metric to define what kind of sub-net structure is optimal. 
We select two sampled non-uniform sub-nets learned by our proposed method, and compare with the uniform ones with the same model size, as shown in~\cref{fig:sturc}. It provides some heuristic thinking for pruning and Neural Architecture Search (NAS) exploration. Our sampled non-uniform structure all have better accuracy compared with uniform sub-nets with the same model size. We summarize the main properties of the sampled non-uniform structures across different network typologies as below: 

\begin{figure}[ht]
  \centering
  \includegraphics 
      [width=1.0  \linewidth]{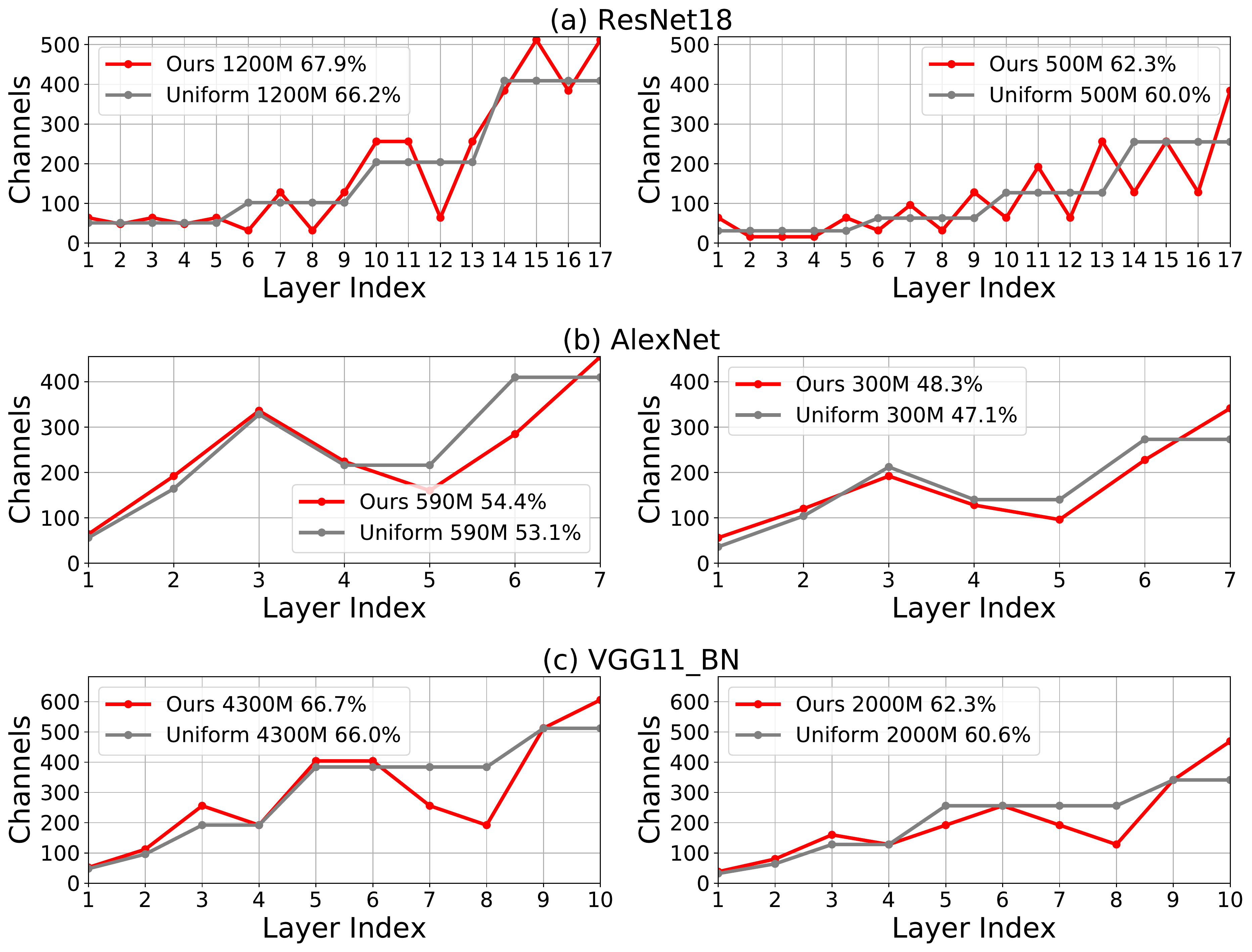}\\
\caption{The sub-net structures with the corresponding accuracy of the proposed progressive searching method on ImageNet dataset for (a) ResNet18, (b) AlexNet, and (c) VGG11-BN. Note that, the fully connected layers are scaled by 9 and 6 for AlexNet  (i.e. the $6_\textrm{th}$ and $7_\textrm{th}$ layers) and VGG11-BN (i.e. the $9_\textrm{th}$ and $10_\textrm{th}$ layers) respectively.}
\label{fig:sturc}
\end{figure}

\begin{itemize}
\item Comparing with uniform structures, we observe that all these three DNNs, i.e., ResNet18, AlexNet and VGG11-BN, have larger number of channels in the first and last layers. It aligns with the general conclusion from many prior works that these two layers are very important in the performance of overall network, which typically need more channels to extract sufficient features or accurately classify into correct groups. 
    
\item ResNet consists of one single convolutional layer followed by several convolutional blocks and a fully connected layer sequentially. Each block includes two convolutional layers and an identity shortcut connection for ResNet18. 
For the blocks starting from the $2_\textrm{nd}$ layer as shown in~\cref{fig:sturc}(a), we observe that the sampled first convolutional layer is smaller and the second one is larger than the uniform structures with the same model size. It might because the second layer receives features from both previous layer and the skip connection, thus requireing large \#channels to avoid information bottleneck. 
  
\item AlexNet is a single-path structure, which includes several convolutional layers and two fully connected layers.
We observe that the sampled sub-nets have less number of channels in the last convolutional layer (i.e. the $5_\textrm{th}$ layer in~\cref{fig:sturc}(b)). This is because the input channel of the first fully connected layer of full size AlexNet is extremely large (i.e. $512\times6\times6$), indicating very high redundancy. Similar phenomena can also be observed in VGG. 

\end{itemize}

\begin{figure}[ht]
  \centering
\includegraphics[width=\linewidth]{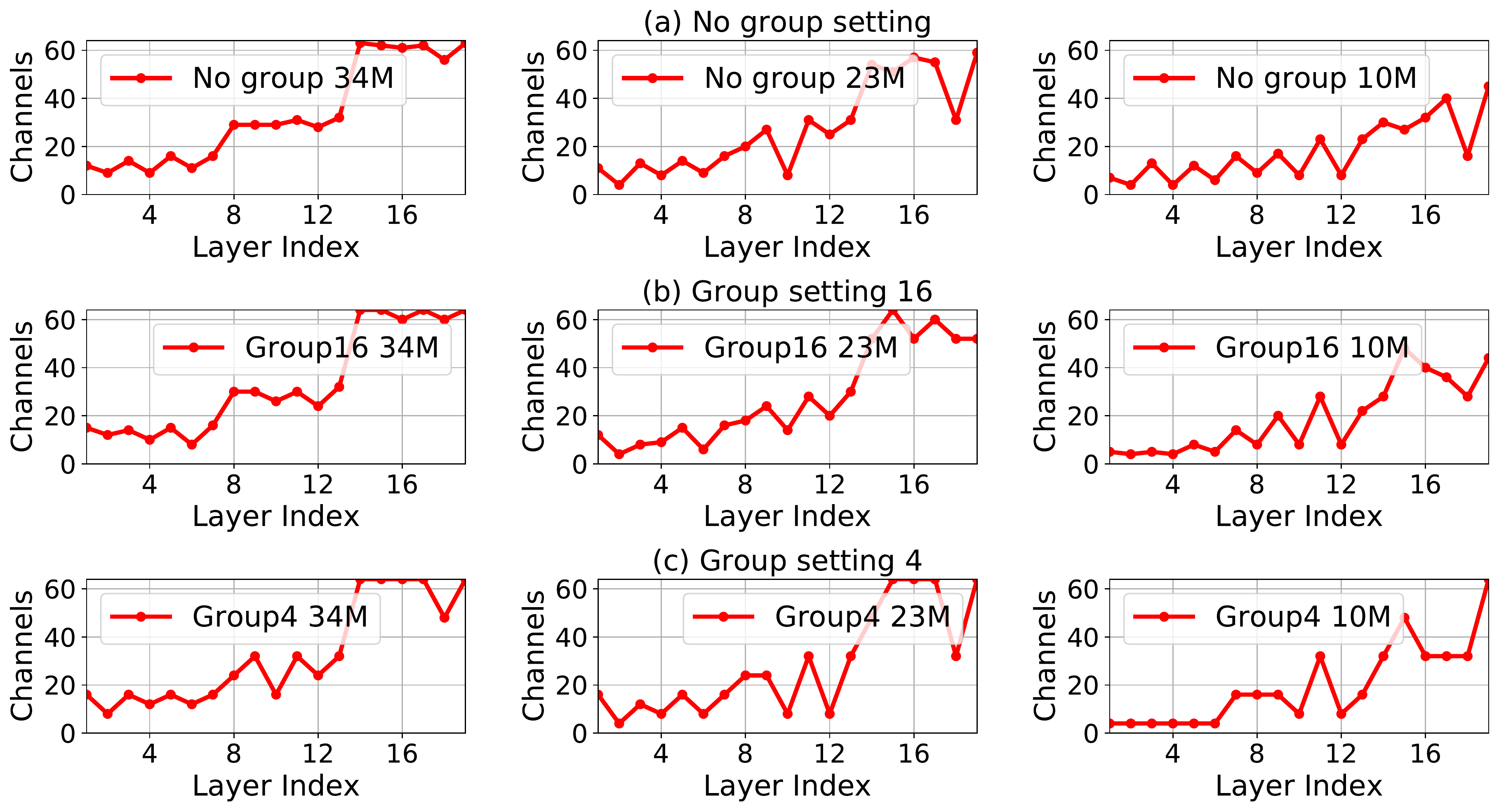}\\
\caption{The sub-net structures with three model sizes for different channel group setting on ResNet20: (Top) No group, (Middle) Group 16 and (bottom) Group 4}
\label{fig:visual_group}
\end{figure}

In addition, we also study how the group channel setting will influence sub-net structures as shown in~\cref{fig:visual_group}. Larger group setting creates fine-grained non-uniform structures since it prunes less channel numbers in each searching iteration. Note that, \textit{No group} setting means only a single channel is pruned in each iteration, which creates the best fine-grained structure. From the experiment results, we observe that the \textit{Group 4} settings, which has the most coarse-grained structure, but with smallest searching cost, still keeps similar characteristics of non-uniform structure, namely larger channel numbers in the layers where the inputs from both previous layer and skip connection layer, while the next connected layer has smaller channel number. For example, the $11_\textrm{th}$ always has much larger channel numbers than layer $12_\textrm{th}$ for all settings. Considering such consistent property in our sub-net searching, it explains why the \textit{Group 4} setting could still achieve very similar performance with other larger group settings as discussed in~\cref{fig:abs_group}, while requiring least searching time. It also supports our claim that the proposed trainable noise ranking is a fast and accurate method to indicate the sensitivity or importance of channels/ groups, thus could be utilized to quickly sample non-uniform sub-nets. 

\subsubsection{CPU and GPU performance}

\begin{figure}[t]
  \centering
  \includegraphics[width=1.0 \linewidth]{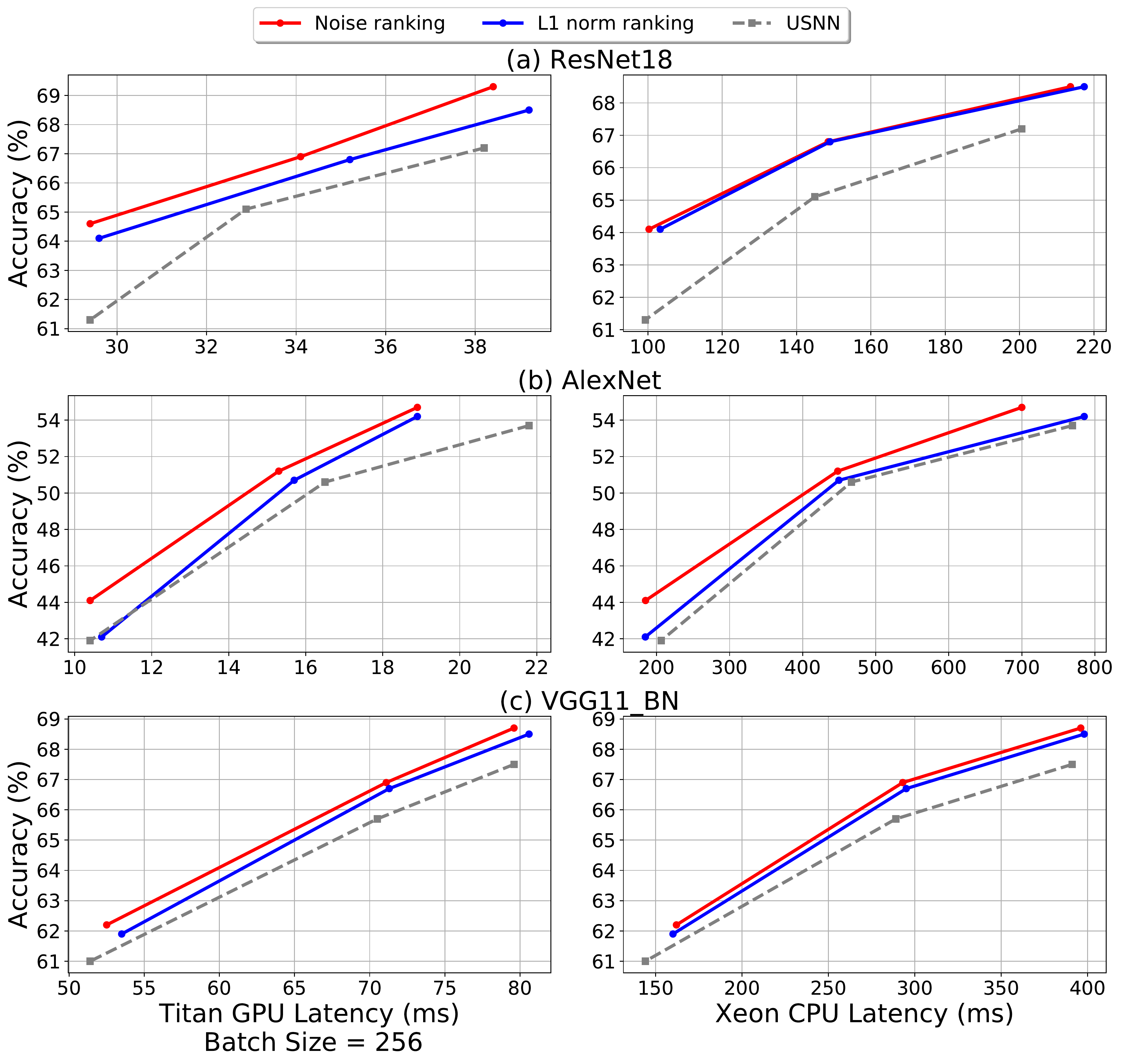}\\
 	\caption{Run-time tuning between latency and accuracy in Titan GPU and Xeon CPU, for (top) ResNet-18, (middle)AlexNet, (bottom) VGG11 on ImageNet}
\label{fig:latency_imagenet}
\end{figure}
We deploy the constructed dynamic inference model based on different network structures in Nvidia Titan-Xp GPU and Intel Xeon CPU as shown in~\cref{fig:latency_imagenet}. It can be seen that our proposed trainable noise ranking method enables run-time tuning between accuracy and latency. It also outperforms US-Net by a large degree in both CPU and GPU execution, showing better accuracy with the same latency. In addition, comparing with $\normlone$-norm ranking, our trainable noise ranking could also achieve better accuracy under same latency.

\section{Conclusion}
In this work, we target to construct a dynamic DNN structure that is able to adjust its inference structure on-the-fly within a group of sub-nets, through a novel proposed trainable noise ranking based sub-net progressive searching method. Extensive experiments on CIFAR-10 and ImageNet datasets indicate that our method could achieve state-of-the-art performance. Beyond that, the constructed dynamic network is deployed to Nvidia Titan GPU and Intel Xeon CPU to demonstrate its dynamic trade-off between accuracy and latency.

\clearpage
\small{
\bibliography{main_arxiv.bib}
}

\clearpage
\appendix

\section{Appendix}
\subsection{Algorithm of Constructing Dynamic Network with Progressive Searching}
The details of proposed progressive sub-network searching for dynamic inference is specified in~\cref{algo:pro_samp}.

\setlength{\textfloatsep}{0pt}
\begin{algorithm}[h]
\caption{The proposed progressive subnet searching for dynamic inference}\label{algo:pro_samp}
\begin{algorithmic}[1]
\Require{Given a pre-trained model $\mW$ with $L$ layers, the defined group channel $G$, the model size constraint $M'$ (e.g. $0.25\times$ ), the accuracy threshold $\gamma$, and noise variance $\bm{\beta}$.}
\State $\hat{Acc} = 0$
\State $subnets_{pool} = []$
\State i = 0
\State $M_i$ = ModelsizeMeasurment($\mW$)
\While{$M_i > M'$}
    \For{$j\gets0, L$}
        \State $\mW'_{ij}$ =  TrainableNoiseRranking($\mW_{ij}$, $G$, $\beta_{ij}$)
        \State $Acc$ = $ACC_{val}(f(\mW'_{ij}))$ 
        \If{$Acc > \hat{Acc}$ 
            \State $\hat{Acc} = Acc$}
            \State $\mW^\star = \mW_{ij}$
        \EndIf
    \EndFor
    \State $M_{i+1}$ = ModelsizeMeasurment($\mW^\star$)
    \State $subnets_{pool}.append(\mW^\star)$
    \If{$Acc < \gamma$}
    \State Fine-tuning($\mW^\star$)
    \EndIf
    \State i += 1
\EndWhile
\State $subnets_{pool}'$ = SubnetsReselection($subnets_{pool}$)
\State Multi-termTraining($subnets_{pool}'$)
\end{algorithmic}
\end{algorithm}

\begin{table}[h]
\caption{Trade-off between accuracy and FLOPs for ResNet18, AlexNet and VGG11-BN on ImageNet. Quantative numbers of \cref{fig:main_exp} in main manuscript}
\scalebox{0.8}{
\begin{tabular}{ccccccc}
\hline
\multirow{2}{*}{Network} & \multicolumn{2}{l}{US-Net} & \multicolumn{2}{l}{L1 Norm} & \multicolumn{2}{l}{Weight Noise} \\ \cline{2-7} 
 & Acc & FLOPs & Acc & FLOPs & Acc & FLOPs \\ \hline
\multirow{10}{*}{ResNet18} & 68.9 & 1818 & 69.7 & 1781 & \textbf{69.9} & 1781 \\ \cline{2-7} 
 & 67.9 & 1605 & 69.5 & 1622 & \textbf{69.7} & 1607 \\ \cline{2-7} 
 & 66.9 & 1311 & 68.3 & 1304 & \textbf{68.5} & 1289 \\ \cline{2-7} 
 & 65.3 & 1005 & 66.8 & 971 & \textbf{66.9} & 971 \\ \cline{2-7} 
 & 63.6 & 886 & 66.3 & 827 & \textbf{66.7} & 839 \\ \cline{2-7} 
 & 61.5 & 671 & 64.1 & 613 & \textbf{65.6} & 598 \\ \cline{2-7} 
 & 59.4 & 574 & 62.7 & 547 & \textbf{62.8} & 544 \\ \cline{2-7} 
 & 57.6 & 457 & \textbf{62.6} & 491 & 62.0 & 481 \\ \cline{2-7} 
 & 55.9 & 378 & \textbf{58.4} & 384 & 57.6 & 371 \\ \cline{2-7} 
 & 51.8 & 186 & \textbf{54.0} & 186 & 53.5 & 186 \\ \hline \hline
 \multirow{10}{*}{AlexNet} & 55.7 & 715 & 55.45 & 715 & \textbf{56.01} & 715 \\ \cline{2-7} 
 & 54.55 & 669 & 55.30 & 662 & \textbf{55.34} & 643 \\ \cline{2-7} 
 & 52.73 & 543 & 53.01 & 492 & \textbf{53.59} & 508 \\ \cline{2-7} 
 & 51.71 & 477 & 52.01 & 433 & \textbf{52.93} & 469 \\ \cline{2-7} 
 & 49.89 & 407 & 50.33 & 369 & \textbf{51.41} & 387 \\ \cline{2-7} 
 & 47.14 & 313 & 46.67 & 270 & \textbf{47.33} & 279 \\ \cline{2-7} 
 & 45.77 & 265 & \textbf{44.43} & 201 & 44.20 & 208 \\ \cline{2-7} 
 & 42.92 & 218 & 42.81 & 160 & \textbf{43.01} & 160 \\ \cline{2-7} 
 & 37.53 & 113 & 40.71 & 113 & \textbf{41.30} & 113 \\ \hline \hline
 \multirow{8}{*}{VGG11-BN} & 68.71 & 7214 & 69.06 & 7214 & \textbf{69.06} & 7214 \\ \cline{2-7} 
 & 67.39 & 5851 & 68.11 & 5743 & \textbf{68.31} & 5575 \\ \cline{2-7} 
 & 66.25 & 4651 & 66.80 & 4661 & \textbf{67.0} & 4469 \\ \cline{2-7} 
 & 64.81 & 3615 & 65.37 & 3583 & \textbf{65.57} & 3369 \\ \cline{2-7} 
 & 62.94 & 2746 & 63.87 & 2775 & \textbf{64.37} & 2825 \\ \cline{2-7} 
 & 60.49 & 1902 & 61.21 & 1914 & \textbf{62.18} & 2028 \\ \cline{2-7} 
 & 57.72 & 1027 & 60.11 & 1439 & \textbf{58.35} & 1027 \\ \cline{2-7} 
 & 54.22 & 787 & 54.15 & 787 & \textbf{56.8} & 787 \\ \cline{2-7} \hline 
\end{tabular}}
\label{tab:main_result}
\end{table}

\subsection{Trainable Noise Ranking as a Typical Network Pruning Method}
To further show that our proposed trainable noise ranking method can sample high quality subnet structure, we select one sampled subnet, and retrain it as a new fixed pruned model to compare with other popular channel pruning methods. As shown in \cref{tab:prune_resnet}, we could achieve state-of-the-art performance comparing with other recent network pruning methods.

\begin{table}[h]
\caption{Comparing trainable noise ranking as a pruning method with other popular methods in ImageNet dataset}
\vspace{-1em}
\begin{adjustbox}{width=\columnwidth,center}
\begin{tabular}{llllll}
\hline
\multicolumn{1}{c}{\multirow{2}{*}{Model}} & \multirow{2}{*}{Method} & \multicolumn{2}{l}{Top-1} & \multirow{2}{*}{FLOPs} & \multirow{2}{*}{\begin{tabular}[c]{@{}l@{}}Prune\\ Ratio\end{tabular}} \\ \cline{3-4}
\multicolumn{1}{c}{} &  & Prune Acc & Acc Drop &  &  \\ \hline
\multirow{5}{*}{ResNet-18} & LCCL\cite{dong2017more} & 66.33\% & 3.65\% & 1.19E9 & 34.6\% \\ \cline{2-6} 
 & SFP\cite{he2018soft} & 67.10\% & 3.18\% & 1.06E9 & 41.8\% \\ \cline{2-6}
 & FPGM\cite{he2018pruning} & 68.41\% & 1.87\% & 1.06E9 & 41.8\% \\ \cline{2-6}
 & TAS\cite{dong2019network} & 69.15\% & 1.50\% & 1.21E9 & 33.3\% \\ \cline{2-6}
 & Ours & 69.05\% & 1.69\% & 1.21E9 & 33.3\% \\ \hline
\end{tabular}
\label{tab:prune_resnet}
\end{adjustbox}
\end{table}

\end{document}